\RequirePackage{fix-cm}

\documentclass[smallextended]{svjour3}   
\usepackage{times}
\usepackage{enumitem}
\usepackage{tabularx}
\usepackage{latexsym}
\usepackage{graphicx}
\usepackage{url}
\usepackage{float}
\usepackage{breakcites}
\usepackage[justification=centering]{caption}
\smartqed  
\begin{document}

\title{Sentiment Analysis at SEPLN (TASS)-2019: Sentiment Analysis at Tweet level using Deep Learning}


\author{Avishek Garain         \and
        Sainik Kumar Mahata 
}


\institute{Avishek Garain  \at
              Computer Science and Engineering\\
              Jadavpur University, Kolkata \\
              \email{avishekgarain@gmail.com}           
           \and
           Sainik Kumar Mahata \at
              Computer Science and Engineering\\
              Jadavpur University, Kolkata \\
              \email{sainik.mahata@gmail.com}  
              }

\date{Received: date / Accepted: date}

\maketitle

\begin{abstract}
This paper describes the system submitted to "Sentiment Analysis at SEPLN (TASS)-2019" shared task. The task includes sentiment analysis of Spanish tweets, where the tweets are in different dialects spoken in Spain, Peru, Costa Rica, Uruguay and Mexico. The tweets are short (up to 240 characters) and the language is informal, i.e., it contains misspellings, emojis, onomatopeias etc. Sentiment analysis includes classification of the tweets into 4 classes, viz., Positive, Negative, Neutral and None. For preparing the proposed system, we use Deep Learning networks like LSTMs. 
\keywords{BiLSTM \and Regularizers \and Sentiment Analysis \and CuDNNLSTM}
\end{abstract}

\section{Introduction}
\label{intro}
\textbf{S}entiment \textbf{A}nalysis (SA) refers to the use of \textbf{N}atural \textbf{L}anguage \textbf{P}rocessing (NLP) to systematically identify, extract, quantify, and study affective states and subjective information. The Sentiment Analysis at SEPLN (TASS)-2019 \footnote{https://competitions.codalab.org/competitions/21957} was a classification task where it was required to classify a Spanish tweet on basis of its sentiment,into various classes like, Positive, Negative, Neutral and None. It was further divided into two subtasks;the first subtask being monolingual testing of system while the second task being cross-lingual testing of system. However, the task threw some additional challenges. The given tweets involved lack of context, where the number of words were less than 240. Moreover, the tweets were in an informal language and contained multi-linguality. Also, the classification system that would be prepared for the task, needed to be generalized for various test corpora as well.

To solve the task in hand, we built a bidirectional \textbf{L}ong \textbf{S}hort \textbf{T}erm \textbf{M}emory (LSTM) based neural network, for prediction of the sentiments present in the provided dataset. For both the subtasks, our system categorized the instances into \texttt{P}, \texttt{N}, \texttt{NEU} and \texttt{NONE}. 

The rest of the paper has been organized as follows. Section \ref{sec:data} describes the data, on which, the task was performed. The methodology followed is described in Section \ref{sec:methodology}. This is followed by the results and concluding remarks in Section \ref{sec:results} and \ref{sec:conclusion} respectively.

\section{Data}
\label{sec:data}The dataset that was used to train the model was provided by InterTASS\cite{10045_27862}. The data was collected from Twitter and it was retrieved using the Twitter API by searching for keywords and constructions that are often included in various texts of different sentiments. The dataset provided consisted of tweets in their original form along with the corresponding \texttt{P}, \texttt{N}, \texttt{NEU} and \texttt{NONE} labels, as shown in Table~\ref{tab1:meaning}.
\vspace{-1em}
\begin{table}[h]
\centering
\begin{tabular}{|c|c|}
\hline
\textbf{Label} & \textbf{Meaning} \\ \hline
P & Positive \\ \hline
N & Negative \\ \hline
NEU & Neutral \\ \hline
NONE & None \\ \hline
\end{tabular}
\vspace{-1em}
\caption{Labels used in the dataset}
\vspace{-1em}
\label{tab1:meaning}
\end{table}
\vspace{-1em}
The dataset originally comprised of Spanish tweets of various dialects, namely \texttt{ES}(Spain), \texttt{PE}(Peru), \texttt{CR}(Costa Rica), \texttt{UR}(Uruguay) and \texttt{MX}(Mexico). The tweets were also tagged with their respective sentiments. We merged all this data and shuffled them. The resulting dataset had 7,265 sentiment tagged tweets, which were splitted into 5,086 instances of training data and 2,179 instances of development data. Our approach was to convert the tweets into a sequence of words and convert them into word embeddings. We then run a neural-network based algorithm on the processed tweet. Language and label based categorical division of data is given in Table~\ref{tab:tab2},~\ref{tab3:dataset-dist-train},~\ref{tab4:dataset-dist-dev} and \ref{tab5:dataset-dist-tot}.
\vspace{-1em}
\begin{table}[h]
\centering
\begin{tabular}{|c|c|c|}
\hline
\textbf{Label} & \textbf{Train} & \textbf{Development} \\ \hline
Spain & 1125 & 581 \\ \hline
Peru & 966 & 498 \\ \hline
Costa Rica & 777 & 390 \\ \hline
Uruguay & 943 & 486 \\ \hline
Mexico & 989 & 510 \\ \hline
\end{tabular}
\vspace{-1em}
\caption{Training and Development data used for the system.}
\label{tab:tab2}
\end{table}

\vspace{-1em}
\vspace{-1em}
\vspace{-1em}
\begin{table}[H]
\center
\begin{tabular}{|c|c|c|c|c|}
\hline
\bf Value & \bf P & \bf NEU & \bf N & \bf NONE \\ 
\hline
\bf All & 1994 & 710 & 1483 & 898\\
\hline
\end{tabular}
\vspace{-1em}
\caption{Distribution of the labels in the training dataset}
\label{tab3:dataset-dist-train}
\end{table}

\vspace{-1em}
\vspace{-1em}
\vspace{-1em}
\begin{table}[H]
\center
\begin{tabular}{|c|c|c|c|c|}
\hline
\bf Value & \bf P & \bf NEU & \bf N & \bf NONE \\ 
\hline
\bf All & 850 & 297 & 615 & 418 \\
\hline
\end{tabular}
\vspace{-1em}
\caption{Distribution of the labels in the development dataset}
\label{tab4:dataset-dist-dev}
\end{table}

\vspace{-1em}
\vspace{-1em}
\vspace{-1em}
\begin{table}[H]
\center
\begin{tabular}{|c|c|c|c|c|}
\hline
\bf Value & \bf P & \bf NEU & \bf N & \bf NONE \\ 
\hline
\bf All & 2844 & 1007 & 2098 & 1316\\
\hline
\end{tabular}
\vspace{-1em}
\caption{Distribution of the labels in the combined dataset}
\vspace{-1em}
\label{tab5:dataset-dist-tot}
\end{table}
\vspace{-1em}
The provided training and development data were merged and shuffled to create a bigger training set, and we refer to the same as training data in the methodology section.

\section{Methodology}
\label{sec:methodology}
The first stage in our model was to preprocess the tweets. For the preprocessing steps, we took inspiration from the work on Hate Speech against immigrants in Twitter\cite{garain-basu-2019-titans}, part of SemEval2019. The steps used here are built as an advancement of this work. It consisted of the following steps:
\begin{enumerate}[nolistsep]
    \item Removing mentions
    \item Removing URLs
    \item Contracting whitespace
    \item Extracting words from hashtags
\end{enumerate}
The last step (step 5) consists of taking advantage of the Pascal Casing of hashtags (e.g. \texttt{\#TheWallStreet}). A simple regex can extract all words; we ignore a few errors that arise in this procedure. This extraction results in better performance mainly because words in hashtags, to some extent, may convey sentiments of hate. They play an important role during the model-training stage.

We treat each tweet as a sequence of words with interdependence among various words contributing to its meaning. We convert the tweets into one-hot vectors. We also include certain manually extracted features listed below:
\begin{enumerate}
    \item Counts of words with positive sentiment, negative sentiment and neutral sentiment in Spanish
    \item Counts of words with positive sentiment, negative sentiment and neutral sentiment in English
    \item Subjectivity score of the tweet
    \item Number of question marks,Exclamations and full-stops in the tweet
\end{enumerate}
For this, we used SenticNet5\cite{Cambria2018SenticNet5D} for finding sentiment values of individual words after converting the sentences to English via GoogleTrans API. Apart from this, we also used a Spanish Sentiment lexicon for the same. 

The use of BiLSTM networks is a key factor in our model. The work of \cite{Hochreiter:1997:LSM:1246443.1246450} brought a revolutionary change by bringing the concept of memory into usage for sequence based problems. We were guided by the work of \cite{zhang2018detecting} who used a CNN+GRU based approach for a similar task. We use an approach which was influenced by this work to some extent.

We use a bidirectional LSTM based approach to capture information from both the past and future context followed by an Attention layer consisting of initializers and regularizers. 

Our model is a neural-network based model. Initially, the manual feature vectors are appended with the feature vector obtained after converting the processed tweet to one-hot encoding. It is then passed through an embedding layer which transforms the tweet into a 128 length vector. The embedding layer learns the word embeddings from the input tweets. We pass the embeddings through a Batch Normalization layer of dimensions 10 X 128. This is followed by one bidirectional LSTM layer containing 128 units with its dropout and regular dropout set to 0.4 and activation being a sigmoid activation. This is followed by a Bidirectional CuDNNLSTM layer with 64 units for better GPU usage. This is followed by the final output layer of neurons with softmax activation, where, each neuron predicts a label as present in the dataset. 

\begin{figure}[h]
    \centering
    \includegraphics[scale=0.6]{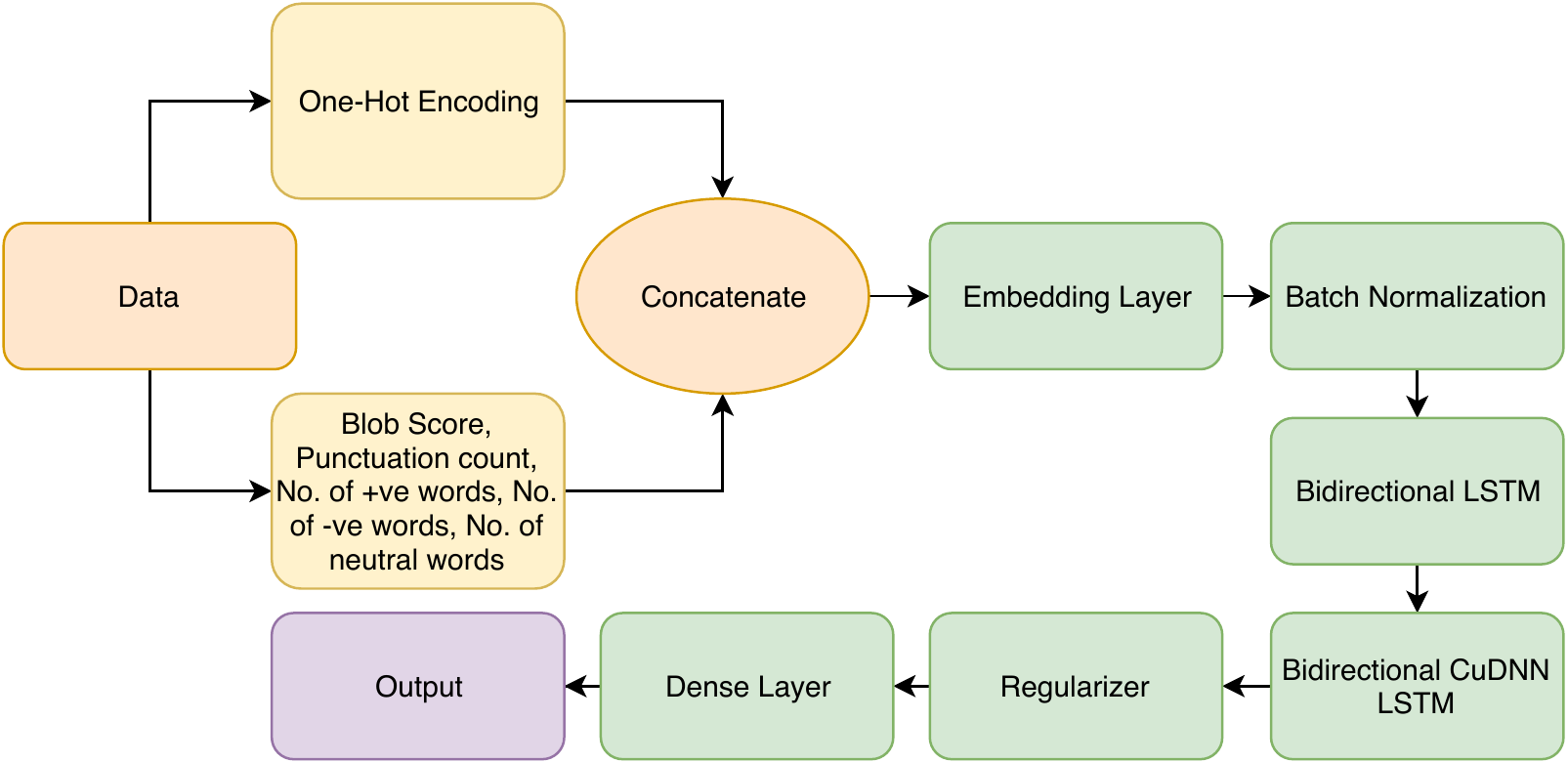}
    \vspace{-1em}
    \caption{Flowchart of working model}
    \label{fig:fig1}
\end{figure}

For both subtasks 1 and 2, we train a model containing 4 neurons for predicting \texttt{P(0/1)}, \texttt{N(0/1)} and \texttt{NEU(0/1)} and \texttt{NONE(0/1)} respectively. After the CuDNNLSTM layer we have added a regularizing layer which is initialized with glorot\textunderscore uniform initializer. This layer has both W\textunderscore regularizers and b\textunderscore regularizers to prevent the model from overfitting. This provides better validation and test results by generalizing the feature learning process. The model is compiled using the Adam optimization algorithm with a learning rate of 0.0005. Categorical-crossentropy is used as the loss function. The working is depicted in Figure \ref{fig:fig1}.
\vspace{-1em}
\vspace{-1em}
\vspace{-1em}
\vspace{-1em}
We note that the dataset is highly skewed in nature. If trained on the entire training dataset without any validation, the model tends to completely overfit to the class with higher frequency as it leads to a higher accuracy score. 

To overcome this problem, we took some measures. Firstly, the training data was split into two parts --- one for training and one for validation comprising 70 \% and 30 \% of the dataset respectively. The training was stopped when two consecutive epochs increased the measured loss function value for the validation set. 

Secondly, class weights were assigned to the different classes present in the data. The weights were approximately chosen to be proportional to the inverse of the respective frequencies of the classes. Intuitively, the model now gives equal weight to the skewed classes and this penalizes tendencies to overfit to the data.

\section{Results}
\label{sec:results}
We participated in subtasks 1 and 2 of Sentiment Analysis at SEPLN (TASS)-2019 and our system ranks first among the competing participants.

We have included the automatically generated tables with our results. The results(rounded off to 3 decimal places) are depicted in Table \ref{tab:with-hastags}, \ref{tab:task1_results} and \ref{tab:task2_results}.
\vspace{-1em}
\begin{table}[H]
\center
\begin{tabular}{|c|c|c|}
\hline
\bf System & \bf Train (\%) & \bf Validation (\%) \\ 
\hline
Without & 63.34 & 48.86 \\
With & 67.18 & 51.98 \\
\hline
\end{tabular}
\vspace{-1em}
\caption{Comparison of development phase accuracies with and without hashtag preprocessing}
\label{tab:with-hastags}
\end{table}
\vspace{-1em}
\vspace{-1em}
\vspace{-1em}
\vspace{-1em}
\begin{table}[H]
\center
\begin{tabular}{|c|c|c|c|c|}
\hline
\textbf{Metric} & \textbf{System} & \textbf{F1} & \textbf{Precision} & \textbf{Recall} \\ \hline
\textbf{CR} & BiLSTM & 0.250 & 0.245 & 0.256 \\ \hline
\textbf{ES} & BiLSTM & 0.261 & 0.265 & 0.258 \\ \hline
\textbf{MX} & BiLSTM & 0.384 & 0.393 & 0.376 \\ \hline
\textbf{PE} & BiLSTM & 0.263 & 0.272 & 0.254 \\ \hline
\textbf{UY} & BiLSTM & 0.218 & 0.240 & 0.201 \\ \hline
\end{tabular}
\vspace{-1em}
\caption{Task 1 Statistics.}
\label{tab:task1_results}
\end{table}
\vspace{-1em}
\vspace{-1em}
\vspace{-1em}
\vspace{-1em}
\begin{table}[H]
\center
\begin{tabular}{|c|c|c|c|c|}
\hline
\textbf{Metric} & \textbf{System} & \textbf{F1} & \textbf{Precision} & \textbf{Recall} \\ \hline
\textbf{CR} & BiLSTM & 0.250 & 0.245 & 0.256 \\ \hline
\textbf{ES} & BiLSTM & 0.261 & 0.265 & 0.258 \\ \hline
\textbf{MX} & BiLSTM & 0.384 & 0.393 & 0.376 \\ \hline
\textbf{PE} & BiLSTM & 0.263 & 0.272 & 0.254 \\ \hline
\textbf{UY} & BiLSTM & 0.218 & 0.240 & 0.201 \\ \hline
\end{tabular}
\vspace{-1em}
\caption{Task 2 Statistics.}
\vspace{-1em}
\label{tab:task2_results}
\end{table}

\section{Conclusion}
\label{sec:conclusion}
In this system report, we have presented a model which performs satisfactorily in the given tasks. The model is based on a simple architecture. There is scope for improvement by including more manually extracted features (like those removed in the preprocessing step) to increase the performance. Another fact is that the model is a constrained system, which may lead to poor results based on the modest size of the data. Related domain knowledge may be exploited to obtain better results. Use of regularizers led to proper generalization of model, henceforth increasing our task submission score.\nocite{*}

\bibliography{tass}
\bibliographystyle{spbasic}

\end{document}